\documentclass{article}

\usepackage{arxiv}

\usepackage[utf8]{inputenc} 
\usepackage[T1]{fontenc}    
\usepackage{hyperref}       
\usepackage{url}            
\usepackage{booktabs}       
\usepackage{amsfonts}       
\usepackage{nicefrac}       
\usepackage{microtype}      
\usepackage{lipsum}		
\usepackage{graphicx}
\usepackage[square,sort,comma,numbers]{natbib}
\usepackage{doi}

\usepackage{multirow}

\title{Machine Learning Constructives and Local Searches for the Travelling Salesman Problem}

\author{ 
    Tommaso~Vitali\thanks{Equal contribution} \\
	Universit\`{a} della Svizzera Italiana\\
    Lugano, Switzerland\\
	\And
	\href{https://orcid.org/0000-0002-8464-1889}{\includegraphics[scale=0.06]{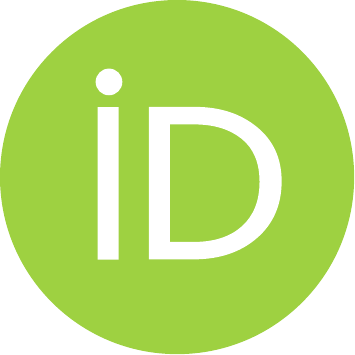}\hspace{1mm}Umberto J.~Mele\footnotemark[1]} \\
	Dalle Molle Institute for Artificial Intelligence\\
	Universit\`{a} della Svizzera Italiana\\
    Lugano, Switzerland\\
	\texttt{umbertojunior.mele@idsia.ch} \\
	\AND
    Luca M.~Gambardella \\
    Dalle Molle Institute for Artificial Intelligence\\
	Universit\`{a} della Svizzera Italiana\\
    Lugano, Switzerland\\
    \And 
    Roberto~Montemanni \\
    Department of Sciences and Methods for Engineering\\
    University of Modena and Reggio Emilia\\
    Reggio Emilia, Italy\\
    \texttt{roberto.montemanni@unimore.it}\\
}

\renewcommand{\shorttitle}{ML-constructives and Local Searches for the TSP}

\hypersetup{
pdftitle={A template for the arxiv style},
pdfsubject={cs.AI, stat.ML},
pdfauthor={Tommaso~Vitali, Umberto J.~Mele, Luca M.~Gambardella, Roberto~Montemanni},
pdfkeywords={Travelling Salesman Problem, Machine Learning, Hybrid Heuristic, Combinatorial Optimization, Artificial Intelligence},
}

\begin{document}
\maketitle

\begin{abstract}
The \emph{ML-Constructive} heuristic is a recently presented method and the first hybrid method capable of scaling up to real scale traveling salesman problems. It combines machine learning techniques and classic optimization techniques.
In this paper we present improvements to the computational weight of the original deep learning model. In addition, as simpler models reduce the execution time, the possibility of adding a local-search phase is explored to further improve performance.
Experimental results corroborate the quality of the proposed improvements.
\end{abstract}

\keywords{Travelling Salesman Problem \and Machine Learning \and Hybrid Heuristic \and Combinatorial Optimization \and Artificial Intelligence}

\section{Introduction} \label{introduction}

The Travelling Salesman Problem (TSP) is one of the most investigated problems in the Combinatorial Optimization (CO) field. This is partly due to the fact that it belongs to the set of NP-Hard problems, which makes it particularly challenging. 
Moreover, the many practical problems that can be reduced to this -- such as in Ratnesh \emph{et al.} \cite{ratnesh} where models of the TSP are presented to be used in the manufacture of microchips -- make it even more attractive.
At the same time, the full potentials of Machine Learning (ML) and Deep Learning (DL)  techniques are becoming increasingly recognized in the CO field \cite{survey}.

Mele et al. \cite{mele:gambardella:montemanni} recently introduced \textit{ML-Constructive}, a promising constructive approach that computes fast solutions in two separate phases.
The first phase uses ML to create a sub-solution with the most reliable edges.
Whereas, the second phase employs a classic heuristic to complete the tour.
Here we introduce an extension to the original idea to enhance the performance of the \textit{ML-Constructive} algorithm. 

\vspace{5 pt}

In \autoref{introduction} we formally state the Travelling Salesman Problem, and present a brief literature review. A high-level description of the plain method is presented in \autoref{method}. In \autoref{improvements} we present all the changes we made to improve the algorithm. Finally, in \autoref{results} the results of the new approach we propose are shown and discussed.

\subsection{The Travelling Salesman Problem} \label{tsp}

Let consider the complete graph $G = (V, E)$, where $V = \{1, ..., n\}$ is a set of $|V| = n$ nodes, and $E = \{e_{ij} : i,j \in V \; \textrm{with} \; i \neq j\}$ is a set of edges connecting nodes to each other. Also, let $c_{ij}$ be the cost for edge $e_{ij}$ connecting node $i$ to node $j$. The objective of the Travelling Salesman Problem is to find the shortest possible tour that visits each node exactly once, and then gets back to the first node \cite{applegate}.
The largest TSP instance ($n = 85900$) solved optimally required more than 136 years of CPU time; the computation has been carried out with the fast Concorde solver \cite{applegate}. 
The NP-hard nature of the problem makes it fundamental the development of algorithms that compute approximate solutions with a good confidence even on large instances.

\vspace{5 pt}

An effective way to heuristically reduce the complexity of a TSP is to consider only subsets of edges when building a feasible tour. 
A Candidate List $CL_i$ for node $i$ is defined as the set of edges 
that contains the most likely edges to be part of the optimal tour. 
There exist different methods to create candidate lists, the simplest one of which is to consider only the edges connecting the $k$ closest nodes to each node $i$. 

\subsection{A Brief Literature Review} \label{literature-review}
It is well known that an efficient way to solve large Combinatorial Optimization problems is to employ the \emph{Divide-and-Conquer} paradigm.
Such a paradigm is promising for addressing CO problems with Machine Learning as well.
Since ML models suffer from a intrinsic generalization problem trying to scale up to large instances \cite{generalize} due to well-known ML limits (e.g. imbalanced training) \cite{ml_flaws}.

Valuable surveys describing recent approaches using ML to generate solutions for CO problems are in \cite{mele_survey, survey}.
Different approaches suggesting DL networks to solve the TSP with end-to-end methodology have been presented among which the studies carried out by Miki \emph{et al.} \cite{image_for_TSP}, Kool \emph{et al.} \cite{attention_for_VRP} and Mele \emph{et al.} \cite{mele_rl}.

The best proposal at the moment is the \emph{ML-Constructive} heuristic \cite{mele:gambardella:montemanni}, which focuses on the development of an efficient interaction between Machine Learning and Combinatorial Optimization techniques. 
It uses candidate lists (CLs) as input to the ML model, and is able to scale up with satisfactory results.
Other approaches attempting to solve the scalability issue were introduce by Fu \emph{et al.} \cite{generalize} and by Fitzpatrick \emph{et al.} \cite{ml_for_cl}, where Machine Learning is used to construct CLs and then classical heuristics for the tour construction are applied.

\section{The Original ML-Constructive Heuristic} \label{method}
The \emph{ML-Constructive} heuristic is a constructive hybrid algorithm composed by two phases. 
The first phase exploits Machine Learning's ability in detecting specific patterns to create an initial partial solution.
This solution comprises the edges most likely to be part of an optimal tour according to the ML learnt patterns.
The second phase instead uses a well-known heuristic to complete the solution.
In fact, some difficulties may arise with ML where data is not adequate, further details can be found in Mele \emph{et al.} \cite{mele:gambardella:montemanni}. 

\vspace{5 pt}

In order to initialize the problem, reduce the search space and create valid inputs for the Machine Learning model, \emph{ML-Constructive} initially computes a candidate list for each node. 
Then, a list $L_P$ of promising edges is created, such that it contains all the edges connecting the closest two vertices for each CL.
The Machine Learning is in charge of checking when an edge $l \in L_P$ has to be used or not for the solution.
To do so, it learns the probability that these edges 
have of being optimal by considering just the CL of the considered $l$ edge as input.
Initially the insertion feasibility of edge $l$ is checked considering the current partial solution, then the ML predicts the probability of $l$ being an optimal edge. 
If such probability is greater than a certain threshold, the edge will be inserted in the current partial solution.

The order in which these requests are tackled is a fundamental choice.
In \emph{ML-Constructive}, the list $L_P$ is sorted according to the positions in the CL and the non-decreasing cost values.
The edges connecting the nearest node in the CL are placed before, then those which are second closest follow.

\vspace{5 pt}

To complete the tour obtained during the Machine Learning phase, the Clarke-Wright (CW) heuristic \cite{clarke:wright} was used. 
Note that no change is made to the edges inserted during the first phase.

\newpage

\section{Improvements to ML-Constructive} \label{improvements}

The original algorithm uses a ResNet architecture \cite{resnet} to confirm the addition of an edge in the solution.
Such an architecture carries a high computational cost we would like to avoid.
Our first contribution is to attempt to reduce it by replacing the ResNet with a different ML model.
Several alternative ML models were examined.

\vspace{5 pt}

In addition, since \emph{ML-Constructive} has some shortcomings in the heuristic part too, a different CL constructor and a third phase are introduced as well.
The new CL constructor exploits the Delaunay triangolarization \cite{delaunay} to speed up the creation of the lists.
The third phase instead increases the quality of the complete solution by introducing a local search on the most uncertain edges since the CW solution can be largely improved.
We point out that the second phase is kept unchanged here from the original algorithm.
As shown by Mele \emph{et al.} \cite{mele:gambardella:montemanni}, even when the \emph{ML-Constructive} is able to predict all the optimal edges in $L_P$, sometimes it does not reach the complete optimal solution in the end.

\subsection{First Phase: Machine Learning Models}
To find a ML model that works accurately and in a short time, several ML models were tried out and tested.
Their performances in terms of predictions quality and tour construction are shown in \autoref{tab:ml-results} and \ref{tab:mlg-results}, respectively.
Five-thousand instances were randomly generated to train these models, with $100 \leq n \leq 1000$. 
The points were sampled in the unit side square, and the optimal solutions were computed with the Concorde solver \cite{applegate}.
The cost between each node in the CL$_i$ of node $i$ were employed as input of the ML, where the cost is the euclidean distance between vertices.
In addition, it is also provided a vector indicating whether that edge is in the current partial solution or not. 
Given such vector of dimension $(k + 1) \cdot k$, the ML model is asked to predict if the first or second neighbor in the CL$_i$ of node $i$ is optimal.

With the aim of preserving a consistent balance between training and testing, each CL$_i$ in the training set were filtered according to the partial optimal solution found using \emph{ML-Constructive} constraints and iterations \cite{mele:gambardella:montemanni}.

\begin{table}[t!]
\scriptsize
\centering
\caption{\label{tab:ml-results}Performance of ML models.}
\begin{tabular}{ccccccc}
\hline
\textbf{Dataset}                                                             & \textbf{Model}     & \textbf{Accuracy} & \textbf{\begin{tabular}[c]{@{}c@{}}Balanced\\ Accuracy\end{tabular}} & \textbf{Precision} & \textbf{TPR} & \textbf{FPR} \\ \hline
\multirow{6}{*}{\textbf{\begin{tabular}[c]{@{}c@{}}1st\\ edge\end{tabular}}} & \textbf{Baseline \cite{mele:gambardella:montemanni}}  & 0,782             & 0,499                                                                & 0,867              & 0,885        & 0,886        \\
                                                                             & \textbf{Linear \cite{linear}}    & 0,490             & 0,669                                                                & 0,970              & 0,424        & 0,086        \\
                                                                             & \textbf{Linear US \cite{linear,under_sampling}} & 0,436             & 0,650                                                                & 0,975              & 0,359        & 0,059        \\
                                                                             & \textbf{ResNet  \cite{mele:gambardella:montemanni}}    & 0,715             & 0,694                                                                & 0,915              & 0,762        & 0,339        \\
                                                                             & \textbf{SVM \cite{linear_SVM}}       & 0,500             & 0,664                                                                & 0,959              & 0,427        & 0,100        \\
                                                                             & \textbf{Ensemble \cite{ensemble, xgboost}}  & 0,525             & 0,679                                                                & 0,962              & 0,456        & 0,099        \\ \hline
\multirow{5}{*}{\textbf{\begin{tabular}[c]{@{}c@{}}2nd\\ edge\end{tabular}}} & \textbf{Baseline \cite{mele:gambardella:montemanni}}  & 0,501             & 0,500                                                                & 0,511              & 0,512        & 0,512        \\
                                                                             & \textbf{Linear \cite{linear}}    & 0,560             & 0,620                                                                & 0,839              & 0,341        & 0,101        \\
                                                                             & \textbf{ResNet  \cite{mele:gambardella:montemanni}}    & 0,504             & 0,538                                                                & 0,816              & 0,104        & 0,028        \\
                                                                             & \textbf{SVM \cite{linear_SVM}}       & 0,458             & 0,547                                                                & 0,763              & 0,198        & 0,104        \\
                                                                             & \textbf{Ensemble \cite{ensemble, xgboost}}  & 0,411             & 0,514                                                                & 0,722              & 0,075        & 0,047       
\end{tabular}
\end{table}

\vspace{5 pt}

To accomplish the task several approaches were engaged: the baseline predictor which randomly predicts using the empirical probabilities of the CL positions \cite{mele:gambardella:montemanni}, the same ResNet architecture introduced by Mele \emph{et al.} \cite{mele:gambardella:montemanni}, a linear classifier \cite{linear}, a linear SVM \cite{linear_SVM}, and finally an Ensemble \cite{ensemble} voting classifier including also an XGBoost \cite{xgboost}. The latter shows the best performance on the test set.
Since the first edge occurrence is quite over-represented we applied an under-sampling technique as well \cite{under_sampling}.
More details on the training settings can be found in the \emph{online} compendium\footnote{All the code for the experiments, the data creation, the training and testing, along with the \emph{online} Compendium can be found in the GitHub repository at \url{https://github.com/tommivitali/ML-Constructive_LS}.}.

\vspace{5 pt}

Several classic metrics are shown in \autoref{tab:ml-results}, an higher True Positive Rate (TPR) and a lower False Positive Rate (FPR) are preferable \cite{mele:gambardella:montemanni}.
The \emph{ML-Constructive} was tested on 54 TSPLIB instances \cite{reinelt}.

\subsection{Third Phase: Local Search} \label{2opt}
The \emph{ML-Constructive} provides good approximated tours, which can however still be improved.
These tours have some flaws, since it is possible to get some crossing edges in them. 
To obtain better solutions we extend the heuristic with a further step, which employs a 2-opt local search \cite{local_search}.
Generally, such local search compares every possible couple of edges.
However, since we are pretty confident about what has been done in the first phase, here we try to improve only the edges obtained during the second phase.
The edges that have been inserted by the ML models (first phase)
will not be modified. 




\section{Results} \label{results}

To compare the results obtained by the original version of \textit{ML-Constructive} \cite{mele:gambardella:montemanni} and what it is proposed in this work, experiments were carried out on the same 54 instances selected by Mele \emph{et al.} 
\cite{mele:gambardella:montemanni} 
from the TSPLIB library \cite{reinelt}. 
The size of the instances varies between 100 and 1748. A brief recap of the results -- with the heuristic executed using several ML models in the first phase -- is shown in \autoref{tab:mlg-results}. 
A more detailed version of this table can be found in the \emph{online} compendium, where the results of each instance are shown and discussed.

The first column \textit{B} is the baseline, while \textit{NN} confirms an edge if it connects the nearest node in the CL.
The other columns show the performance using some ML models; the column ML-C shows the results of the execution of the original \emph{ML-Constructive} algorithm. 
On the two columns indicated by ``LS" is performed also the 2-opt local search as a third phase of the algorithm. Clearly, this leads overall to better performance: 2-opt moves are applied only if they bring a better tour length.

The introduction of new ML models has brought an improvement in terms of computational burden for the first phase.
In terms of quality, the use of SVM also brought an improvement (not significant) compared to ML-C.
The result is attractive as it also leads to an improvement in speed of about 4x.
More work must be carried out to improve the accuracy of the Machine Learning decision-taker, and
we noticed that keeping low FPR is preferable to having high TPR.

\vspace{5 pt}

The local research introduced shows an improvement in terms of solution quality as well, although more effort is required to bring the gap of the tour established after the local search to zero.
Overall, the changes we made led to better performance with respect to the original \emph{ML-Constructive}, apart from a few particular instances.
The promising results obtained by the ``optimal'' ML policy (OPT) suggest that there's room for improvement along this direction. 
Recall that the OPT policy is derived on the assumption that the ML decision-taker can correctly predict all the optimal edges in $L_P$ without making any mistakes.


\begin{table}[!t]
\scriptsize
\centering
\caption{\label{tab:mlg-results}Brief statistics computed on the results of the modified ML-Constructive heuristic executed on 54 TSPLIB instances. Each column refers to a different ML model.} 
\begin{tabular}{cccccccccc}
\hline
              & \textbf{\begin{tabular}[c]{@{}c@{}}B\\ \cite{mele:gambardella:montemanni}\end{tabular}}
              & \textbf{\begin{tabular}[c]{@{}c@{}}NN\\ \cite{mele:gambardella:montemanni}\end{tabular}} 
              & \textbf{\begin{tabular}[c]{@{}c@{}}LinU\\ \cite{linear}\end{tabular}} 
              & \textbf{\begin{tabular}[c]{@{}c@{}}SVM\\ \cite{linear_SVM}\end{tabular}} 
              & \textbf{\begin{tabular}[c]{@{}c@{}}ENS\\ \cite{ensemble}\end{tabular}}
              & \textbf{\begin{tabular}[c]{@{}c@{}}ML-C\\ \cite{mele:gambardella:montemanni}\end{tabular}} 
              & \hspace{6 pt} \textbf{SVM+LS} & \textbf{\begin{tabular}[c]{@{}c@{}}OPT\\ \cite{mele:gambardella:montemanni}\end{tabular}} & \textbf{OPT+LS} \\ \hline
\textbf{avg}  & 12,66      & 8,82        & 9,46          & 7,98         & 9,20         & 8,03          & 5,56            & 4,47         & 2,96            \\
\textbf{std}  & 2,99       & 1,95        & 2,75          & 1,84         & 2,57         & 1,87          & 1,61            & 2,45         & 2,35            \\
\textbf{best} & 0/54       & 9/54        & 6/54          & 10/54        & 4/54         & 25/54         & 48/54           & 49/54        & 47/54           \\
\textbf{time} & 0,932      & 1,909       & 3,286         & 2,605        & 5,559        & 9,822         & 631,665         & 0,483        & 36,632         
\end{tabular}
\end{table}

\section*{Acknowledgement}
Umberto Junior Mele was supported by the Swiss National Science Foundation
through grants 200020-182360: ``Machine learning and sampling-based metaheuristics for stochastic vehicle routing problems''.







\end{document}